\pdfoutput=1
\documentclass[11pt]{article}

\usepackage[]{acl}

\usepackage{times}
\usepackage{latexsym}

\usepackage[T1]{fontenc}
\usepackage[utf8]{inputenc}

\usepackage{microtype}

\usepackage{inconsolata}

\usepackage{graphicx}

\usepackage{xcolor}
\usepackage{booktabs}
\usepackage{amssymb}
\usepackage{amsmath}
\usepackage{multirow}
\usepackage{graphicx}
\usepackage{colortbl}
\usepackage{adjustbox}

\newcommand{\ours}{\textsc{Munch}}
\title{\includegraphics[width=0.03\textwidth]{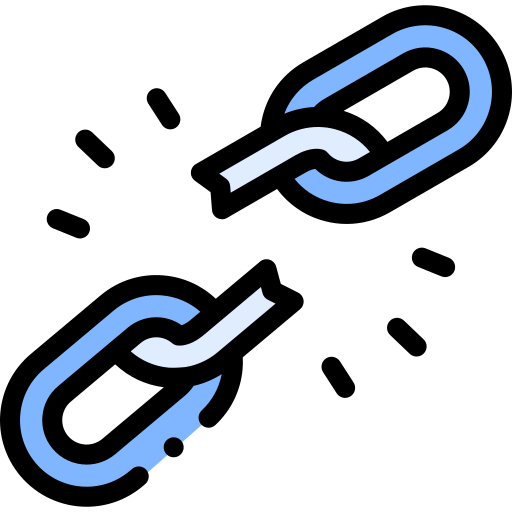} Breaking Chains: Unraveling the Links in\\Multi-Hop Knowledge Unlearning}

\author{Minseok Choi \hspace{0.5cm} ChaeHun Park \hspace{0.5cm} Dohyun Lee \hspace{0.5cm} Jaegul Choo \\
  KAIST AI \\
  \texttt{\{minseok.choi,ddehun,aiclaudev,jchoo\}@kaist.ac.kr} \\
}

\begin{document}
\maketitle
\begin{abstract}

Large language models (LLMs) serve as giant information stores, often including personal or copyrighted data, and retraining them from scratch is not a viable option.
This has led to the development of various fast, approximate unlearning techniques to selectively remove knowledge from LLMs.
Prior research has largely focused on minimizing the probabilities of specific token sequences by reversing the language modeling objective.
However, these methods still leave LLMs vulnerable to adversarial attacks that exploit indirect references.
In this work, we examine the limitations of current unlearning techniques in effectively erasing a particular type of indirect prompt: multi-hop queries.
Our findings reveal that existing methods fail to completely remove multi-hop knowledge when one of the intermediate hops is unlearned.
To address this issue, we propose \ours{}, a simple uncertainty-based approach that breaks down multi-hop queries into subquestions and leverages the uncertainty of the unlearned model in final decision-making.
Empirical results demonstrate the effectiveness of our framework, and \ours{} can be easily integrated with existing unlearning techniques, making it a flexible and useful solution for enhancing unlearning processes.\footnote{To reproduce this work, refer to our code at \url{https://github.com/brightjade/Munch}.}

\end{abstract}

\section{Introduction}

As the volume of data used to train large language models (LLMs) grows exponentially, these models have become vast repositories of information~\cite{carlini2021extracting}.
However, this creates a formidable challenge when specific data from the models need to be removed.
For instance, sensitive information, such as personal or copyrighted data, may unintentionally be included in the training mix, or individuals may exercise their Right to be Forgotten (RTBF)~\cite{rosen2011rtbf} under privacy laws such as the European Union's General Data Protection Regulation (GDPR)~\cite{hoofnagle2019gdpr} or the California Consumer Privacy Act (CCPA)~\cite{pardau2018ccpa} in the United States.
These regulations mandate the removal of personal or protected information from databases, extending to data embedded within machine learning models.
In such cases, model owners must develop mechanisms to safely eliminate specific data while preserving the model's overall functionality.

\begin{figure}
    \centering
    \includegraphics[width=\linewidth]{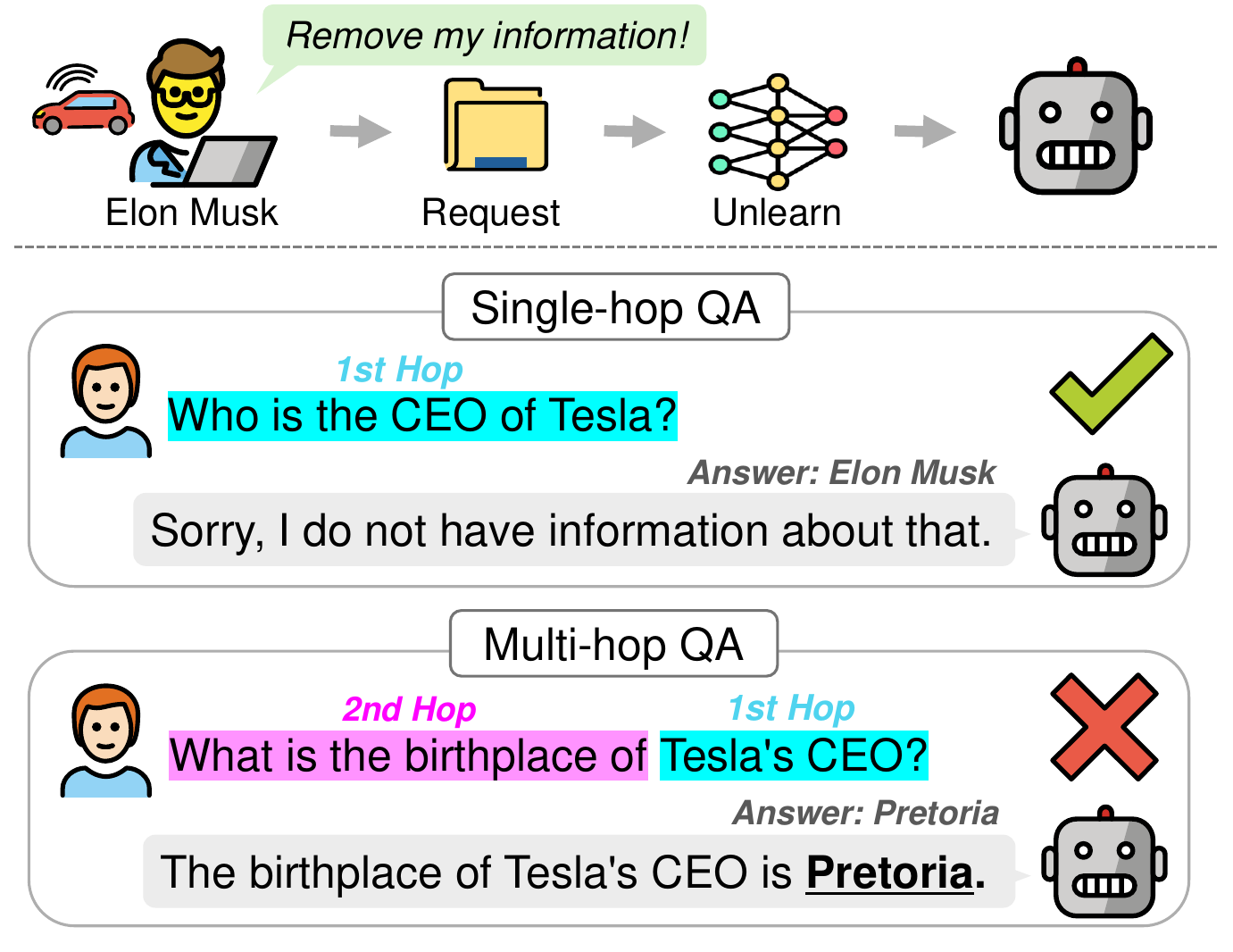}
    \caption{\textbf{Motivation for multi-hop knowledge unlearning.} After Elon Musk (i.e., ``the user'') requests his personal information to be removed from the LLM, existing unlearning methods often succeed in deleting direct, single-hop facts but fail on indirect, multi-hop facts that entail one or a few of the unlearned facts.}
    \label{fig:motivation}
    \vspace{-0.3cm}
\end{figure}

To address these concerns, there has been increasing focus on the field of machine unlearning, which involves removing the influence of specific data points from machine learning models~\cite{cao2015mu}.
Although the need for this task is critical, erasing the effects of certain data on models with billions of parameters is extremely difficult.
The ideal approach is \textit{exact unlearning}, where models are entirely retrained from scratch after excluding the data points that need to be forgotten.
However, this process is computationally intensive and impractical, particularly for LLMs.
As a result, research has shifted towards developing faster \textit{approximate unlearning} techniques.
While machine unlearning has been primarily explored in computer vision~\cite{golatkar2020eternal, golatkar2020forgetting, bourtoule2021sisa, kurmanji2023scrub, fan2024salun}, its prominence is now expanding in NLP due to privacy concerns related to LLMs~\cite{nasr2023scalable, carlini2024stealing}.

Recently, several machine unlearning methods have been introduced in NLP~\cite{jang2023knowledgeunlearning, lee2024pop, zhang2024npo}, with the goal of reversing gradients to prevent LLMs from generating certain sensitive token sequences.
However, these approaches may be vulnerable to adversarial attacks, where specific token sequences are replaced or aliased with alternative sequences.
For example, prompting in low-resource languages has been shown to jailbreak GPT-4~\cite{yong2023lowresource}, and \citet{choi2024crosslingual} demonstrated that current unlearning techniques lack cross-lingual transfer, making LLMs susceptible to such low-resource language exploits.
This leads to an important research question: \textit{``Do current unlearning methods effectively erase multi-hop knowledge when one of the intermediate hops is removed?''}
As illustrated in Figure~\ref{fig:motivation}, consider a scenario where Elon Musk (i.e., ``the user'') requests the removal of his personal information from an LLM.
After unlearning, we expect that direct, single-hop knowledge related to Elon Musk, such as \textit{``Who is the CEO of Tesla?''}, would be deleted.
Additionally, we would expect associated multi-hop knowledge, like \textit{``What is the birthplace of Tesla's CEO?''}, which indirectly references Musk, to also be removed.

In this study, we explore the effectiveness of existing unlearning methods in removing multi-hop knowledge.
We begin by refashioning the widely used multi-hop knowledge editing dataset, MQuAKE~\cite{zhong2023mquake}.
Since we do not need to edit knowledge, we discard edited facts and only consider the original facts for unlearning.
Each example in MQuAKE comprises a multi-hop question (ranging from 2 to 4 hops) that corresponds to a sequence of interconnected facts.
When we unlearn one or more facts within a chain, the model is expected to propagate these changes such that it can no longer answer the associated multi-hop questions.
Our preliminary experiments show that current unlearning methods struggle to forget multi-hop questions when one of the intermediate hops is removed.
For example, ground-truth token sequences could still be extracted from 90.0\% of multi-hop questions after unlearning in the Llama-3.1-8B-Instruct model using NPO~\cite{zhang2024npo}, despite the original extraction success rate of 98.1\% before unlearning.

To achieve more faithful knowledge unlearning, we propose a simple yet effective approach, \ours{}, which significantly outperforms existing approaches in unlearning multi-hop knowledge.
\ours{} first decomposes multi-hop questions into successive subquestions, generates provisional answers, and employs the uncertainty of the unlearned model on the generated outputs as a measure to determine whether to provide a rejective response (e.g., ``I don't know.'') or keep it as is.
Our method capitalizes on the high uncertainty of the unlearned model when dealing with direct, single-hop facts -- an effect stemming from the reversed language modeling objective. By inspecting the decomposed multi-hop questions, we can more easily distinguish between information that needs to be forgotten and information that should be retained.
Empirical results on the modified MQuAKE dataset confirm the efficacy of our approach, and we emphasize that \ours{} is highly practical, requiring no additional training and integrating seamlessly with existing unlearning techniques.
To our knowledge, this is the first work to explore the unlearning of multi-hop knowledge.

\section{Problem Definition}

\subsection{Probing Factual Knowledge in LLMs} \label{sec:probing}

We express a fact as a triple $(s, r, o)$, where $s$ is the subject, $r$ the relation, and $o$ the object.
Following \citet{petroni2019lama}, we define that a pretrained language model possesses specific factual knowledge if it can accurately predict the object $o$ when given the subject $s$ and relation $r$.
For example, if the subject is \textit{Tesla} and the relation is \textit{chief executive officer}, the model should be able to answer the question, \textit{``Who is the CEO of Tesla?''}.
While earlier work primarily focused on cloze-style statements, such as \textit{``The CEO of Tesla is \_\_.''}, using manually written templates, we employ natural language questions to effectively query chat-based models that are becoming widely used.

\subsection{Knowledge Unlearning}

Given a token sequence $\mathbf{x} = \{x\}_{i=1}^T$ from the training dataset $\mathcal{D} = \{\mathbf{x}\}_{i=1}^{N}$, knowledge unlearning aims to safely remove the influence of a specific subset of data $\mathcal{D}_f$ from a trained machine learning model.
The goal is to make the model behave as if this removed data was never used during training, while still maintaining its performance on the remaining dataset.
Typically, the data to be forgotten $\mathcal{D}_f$ is denoted as the \textit{forget set}, and the data to be retained $\mathcal{D}_r$ is referred to as the \textit{retain set}.
For simplicity, we consider the standard case where $\mathcal{D}_f$ and $\mathcal{D}_r$ are mutually exclusive subsets of the entire training dataset, meaning $\mathcal{D}_f \cup \mathcal{D}_r = \mathcal{D}$ and $\mathcal{D}_f \cap \mathcal{D}_r = \varnothing$.
In the context of factual knowledge unlearning, each token sequence $\mathbf{x}$ represents a fact (e.g., \textit{``The CEO of Tesla is Elon Musk.''}), and the objective is to train the model $\pi_\theta$ so that the updated model $\pi_{\theta'} = S(\pi_\theta; \mathcal{D}_f)$ reflects the removal of $\mathcal{D}_f$.
The unlearning function $S$ ensures that the model behaves as if it had only been trained on $\mathcal{D}_r$, effectively forgetting $\mathcal{D}_f$ while preserving its performance on the retained data.

\subsection{Assessing Multi-Hop Queries}

To evaluate the unlearning of multi-hop knowledge, we must first consider a chain of facts $\mathcal{C} = \langle (s_1, r_1, o_1), \ldots, (s_n, r_n, o_n) \rangle$, where the object of the $i^{\text{th}}$ fact also serves as the subject of the next fact in the chain, i.e., $o_i = s_{i+1}$.
Using this chain, we formulate a multi-hop question that starts with the head entity $s_1$ and ends with the tail entity $o_n$.
For instance, consider a chain of two facts: (\textit{Tesla}, \textit{chief executive officer}, \textit{Elon Musk}) and (\textit{Elon Musk}, \textit{place of birth}, \textit{Pretoria}).
This could generate a 2-hop question such as: \textit{``What is the birthplace of Tesla's CEO?''}
When one or more facts from the chain are unlearned, an LLM should adjust its reasoning accordingly, effectively losing the ability to correctly answer the question.
There may be a debate over how many hops should be unlearned, or whether certain multi-hop knowledge should be unlearned at all, as theoretically, the interconnected nature of facts could lead to the unlearning of broader knowledge in the LLM.
For the scope of this study, we focus on the dataset used in our experiments; nevertheless, we hope these discussions inspire further insights into developing more effective and reliable knowledge unlearning methods.

\section{Evaluating Unlearning Approaches on Multi-Hop Question Answering}

\subsection{Dataset}

In this work, we use the MQuAKE benchmark (Multihop Question Answering for Knowledge Editing) introduced by \citet{zhong2023mquake} to evaluate our unlearning approach. This benchmark, designed for knowledge editing tasks, tests whether models can adapt to updates in factual knowledge by modifying their responses to multi-hop queries when individual facts are altered. The benchmark comprises two datasets: \textit{MQuAKE-CF}, which focuses on counterfactual scenarios, and \textit{MQuAKE-T}, which addresses temporal knowledge updates, replacing outdated facts with current information. Both datasets are based on Wikidata and consist of knowledge triplets for single-hop reasoning, as well as multi-hop chains derived from these triplets. Each instance in the benchmark includes: (1) an edit set of single-hop knowledge triplets  $(s, r, o \rightarrow o^*)$, where $o^*$ represents the updated object; (2) a chain of facts $\mathcal{C}$ and its updated version $\mathcal{C}^*$ after knowledge editing; and (3) questions about both the single-hop knowledge and multi-hop chains, before and after the edits.

\begin{table}[t!]
\small
\centering
\resizebox{\columnwidth}{!}{%
\begin{tabular}{lcccc}
\toprule
                            \noalign{\vskip -0.5ex}
                            &   \multirow{2}{*}{\textbf{Forget}}                & \multicolumn{3}{c}{\textbf{Retain}} \\
                            &   & Train& Valid& Test \\ 
                            \noalign{\vskip 0.2ex}
\hline
% \noalign{\vskip 0.5ex}
\multicolumn{5}{l}{\cellcolor[HTML]{EFEFEF}\textit{\textbf{Single-hop}}} \\
\hline
\# of questions                                &1,046  &7,322 &1,046  &1,046 \\
Avg. question length &8.7 &8.6 &8.7 &8.7 \\
% Avg. question length &8.66 &8.59 &8.72 &8.68 \\
% Avg. answer length &9.51 &9.55 &9.55 &9.62 \\
\hline
\multicolumn{5}{l}{\cellcolor[HTML]{EFEFEF}\textit{\textbf{Multi-hop}}} \\ \hline
\# of questions                                &1,036  & - &988  &976 \\
Avg. question length &14.4 &- & 14.5&14.5 \\
% Avg. question length &14.37 &- & 14.47&14.53 \\
% Avg. answer length & -& & 15.14&15.32 \\
Avg. total hops & 2.3&- &2.4 &2.4 \\
% Avg. total hop & 2.32&- &2.35 &2.35 \\
% Avg. UL hop & 1.01 & -& -& -\\
Avg. UL hops & 1.0 & -& -& -\\
\noalign{\vskip -0.3ex}
\bottomrule
\end{tabular}
}
\caption{\textbf{Dataset statistics}. The length of questions denotes the number of words. 
When constructing the retain set for training, we randomly sample the same number of instances as in the forget set. \textit{UL hops} denotes the number of unlearned single-hop facts.}
\label{table:data-stat}
\vspace{-0.3cm}
\end{table}

To adapt this dataset for multi-hop knowledge unlearning, we preprocess the data by: (1) collecting all unique single-hop triplets before the knowledge update; (2) splitting the triplets into a forget set and a retain set at a predefined ratio, with the retain set further divided into training, validation, and test splits; and (3) ensuring that any triplet involved in more than two multi-hop questions is assigned to the retain set's training split. (4) We also ensure that multi-hop questions are appropriately linked to the corresponding single-hop triplets from both the forget and retain sets. If a multi-hop question contains both a forget and retain triplet, it is assigned exclusively to the forget set, ensuring that the final multi-hop forget and retain sets remain mutually exclusive. The number of questions in the single-hop forget set represents 10\% of the total single-hop instances in the MQuAKE dataset.

% \usepackage{graphicx}
\begin{table*}[]
\centering
\small
\resizebox{\textwidth}{!}{%
\begin{tabular}{lcccccccccccc}
\toprule
\multicolumn{1}{l|}{} &
  \multicolumn{3}{c|}{\textbf{Forget Set (Single-Hop)}} &
  \multicolumn{3}{c|}{\textbf{Forget Set (Multi-Hop)}} &
  \multicolumn{3}{c|}{\textbf{Retain Set (Single-Hop)}} &
  \multicolumn{3}{c}{\textbf{Retain Set (Multi-Hop)}} \\
\multicolumn{1}{l|}{} &
  \textbf{PA}($\downarrow$) &
  \textbf{R-L}($\downarrow$) &
  \multicolumn{1}{c|}{\textbf{LM}($\uparrow$)} &
  \textbf{PA}($\downarrow$) &
  \textbf{R-L}($\downarrow$) &
  \multicolumn{1}{c|}{\textbf{LM}($\uparrow$)} &
  \textbf{PA}($\uparrow$) &
  \textbf{R-L}($\uparrow$) &
  \multicolumn{1}{c|}{\textbf{LM}($\downarrow$)} &
  \textbf{PA}($\uparrow$) &
  \textbf{R-L}($\uparrow$) &
  \textbf{LM}($\downarrow$) \\ \\[-2ex]\hline
\multicolumn{13}{l}{\cellcolor[HTML]{EFEFEF}\textit{Llama-3.1-8B-Instruct}} \\ \hline \\[-2ex]
\multicolumn{1}{l|}{Original} &
  99.2 &
  58.1 &
  \multicolumn{1}{c|}{0.6} &
  98.1 &
  29.2 &
  \multicolumn{1}{c|}{1.1} &
  98.9 &
  60.0 &
  \multicolumn{1}{c|}{0.6} &
  98.3 &
  28.0 &
  1.0 \\ \midrule
\multicolumn{1}{l|}{GA\textsuperscript{*}} &
  12.1 &
  0.0 &
  \multicolumn{1}{c|}{146.9} &
  10.3 &
  0.0 &
  \multicolumn{1}{c|}{148.4} &
  12.4 &
  0.0 &
  \multicolumn{1}{c|}{146.4} &
  12.9 &
  0.0 &
  148.2 \\
\multicolumn{1}{l|}{DPO\textsuperscript{*}} &
  30.6 &
  0.8 &
  \multicolumn{1}{c|}{115.1} &
  32.1 &
  1.0 &
  \multicolumn{1}{c|}{113.6} &
  32.0 &
  0.9 &
  \multicolumn{1}{c|}{115.0} &
  32.7 &
  0.5 &
  113.6 \\
\multicolumn{1}{l|}{NPO\textsuperscript{*}} &
  11.0 &
  0.0 &
  \multicolumn{1}{c|}{115.8} &
  21.8 &
  0.0 &
  \multicolumn{1}{c|}{115.1} &
  14.1 &
  0.0 &
  \multicolumn{1}{c|}{115.7} &
  22.1 &
  0.0 &
  115.1 \\ \midrule
\multicolumn{1}{l|}{GA+RT} &
  66.8 &
  49.6 &
  \multicolumn{1}{c|}{\textbf{9.5}} &
  \textbf{88.3} &
  27.4 &
  \multicolumn{1}{c|}{\textbf{5.6}} &
  92.5 &
  71.7 &
  \multicolumn{1}{c|}{4.0} &
  93.0 &
  33.7 &
  4.4 \\
\multicolumn{1}{l|}{DPO+RT} &
  \textbf{65.4} &
  \textbf{36.3} &
  \multicolumn{1}{c|}{6.4} &
  90.3 &
  \textbf{24.0} &
  \multicolumn{1}{c|}{3.4} &
  \textbf{94.6} &
  71.3 &
  \multicolumn{1}{c|}{\textbf{1.9}} &
  \textbf{95.4} &
  33.3 &
  \textbf{2.8} \\
\multicolumn{1}{l|}{NPO+RT} &
  66.8 &
  51.2 &
  \multicolumn{1}{c|}{8.8} &
  90.0 &
  29.1 &
  \multicolumn{1}{c|}{5.6} &
  94.4 &
  \textbf{74.0} &
  \multicolumn{1}{c|}{2.9} &
  \textbf{95.4} &
  \textbf{34.5} &
  3.6 \\ \\[-2ex]\hline \hline
\multicolumn{13}{l}{\cellcolor[HTML]{EFEFEF}\textit{Phi-3.5-Mini-Instruct}} \\ \hline \\[-2ex]
\multicolumn{1}{l|}{Original} &
  89.1 &
  80.4 &
  \multicolumn{1}{c|}{3.7} &
  81.4 &
  52.6 &
  \multicolumn{1}{c|}{3.5} &
  87.7 &
  81.2 &
  \multicolumn{1}{c|}{3.7} &
  81.5 &
  53.8 &
  3.4 \\ \midrule
\multicolumn{1}{l|}{GA\textsuperscript{*}} &
  1.3 &
  0.0 &
  \multicolumn{1}{c|}{198.1} &
  12.3 &
  0.0 &
  \multicolumn{1}{c|}{190.4} &
  4.0 &
  0.0 &
  \multicolumn{1}{c|}{197.4} &
  11.0 &
  0.0 &
  189.9 \\
\multicolumn{1}{l|}{DPO\textsuperscript{*}} &
  9.7 &
  0.0 &
  \multicolumn{1}{c|}{119.0} &
  11.4 &
  0.2 &
  \multicolumn{1}{c|}{116.5} &
  8.3 &
  0.0 &
  \multicolumn{1}{c|}{119.0} &
  11.9 &
  0.1 &
  116.2 \\
\multicolumn{1}{l|}{NPO\textsuperscript{*}} &
  22.4 &
  0.0 &
  \multicolumn{1}{c|}{140.1} &
  24.9 &
  0.0 &
  \multicolumn{1}{c|}{138.3} &
  21.8 &
  0.0 &
  \multicolumn{1}{c|}{140.0} &
  22.1 &
  0.0 &
  138.1 \\ \midrule
\multicolumn{1}{l|}{GA+RT} &
  \textbf{34.9} &
  37.9 &
  \multicolumn{1}{c|}{\textbf{27.9}} &
  \textbf{80.9} &
  26.6 &
  \multicolumn{1}{c|}{\textbf{7.7}} &
  65.8 &
  56.2 &
  \multicolumn{1}{c|}{13.8} &
  85.3 &
  30.8 &
  6.4 \\
\multicolumn{1}{l|}{DPO+RT} &
  38.8 &
  \textbf{6.1} &
  \multicolumn{1}{c|}{12.8} &
  91.1 &
  \textbf{6.7} &
  \multicolumn{1}{c|}{3.3} &
  \textbf{81.1} &
  28.9 &
  \multicolumn{1}{c|}{\textbf{4.2}} &
  \textbf{95.1} &
  9.8 &
  \textbf{2.6} \\
\multicolumn{1}{l|}{NPO+RT} &
  36.6 &
  39.3 &
  \multicolumn{1}{c|}{18.7} &
  88.5 &
  26.9 &
  \multicolumn{1}{c|}{4.1} &
  73.3 &
  \textbf{58.2} &
  \multicolumn{1}{c|}{6.7} &
  90.2 &
  \textbf{31.9} &
  3.3 \\ \bottomrule
\end{tabular}%
}
\caption{Performance comparison of different knowledge unlearning methods after erasing single-hop facts from the forget set in Llama-3.1-8B-Instruct and Phi-3.5-Mini-Instruct models, utilizing 5\% of data as the forget set. \textbf{PA} refers to probing accuracy, indicating the model's success in extracting the targeted knowledge, while \textbf{R-L} represents the ROUGE-L recall score, which measures the alignment between the model's generated outputs and the ground-truth answers. \textbf{LM} denotes the language modeling loss, reflecting the model's unfamiliarity with the token sequences. Models marked with (*) indicate collapse. The best results are highlighted in \textbf{bold}.}
% \caption{Performance of various knowledge unlearning methods after erasing single-hop facts in the forget set from Llama-3.1-8B-Instruct and Phi-3.5-Mini-Instruct. \textbf{PA} denotes the probing accuracy representing the knowledge extraction success rate from the model, and \textbf{R-L} is the ROUGE-L recall score comparing the model's actual generated outputs with the ground-truth answers. \textbf{LM} indicates the language modeling loss, demonstrating how much the token sequences are unfamiliar to the model. (*) indicates collapsed models. The best results are in \textbf{bold}.}
\label{tab:prelim_exp}
\vspace{-0.25cm}
\end{table*}

The statistics for the final dataset are shown in Table \ref{table:data-stat}.  In the single-hop section, the forget set contains 1,046 questions, while the retain set consists of 7,322 training, 1,046 validation, and 1,046 test questions. For multi-hop questions, the forget set includes 1,036 questions, and the retain set has 988 validation and 976 test questions. The average number of total hops is around 2.3 to 2.4. Most of the multi-hop questions in the forget set involve unlearning one single-hop question. 

\subsection{Experimental Setup}

\paragraph{Implementation details}

We built our framework on PyTorch~\cite{paszke2019pytorch} and Hugging Face Transformers~\cite{wolf2020transformers}.
We employed Llama-3.1-8B-Instruct~\cite{dubey2024llama3} and Phi-3.5-Mini-Instruct~\cite{abdin2024phi3} as the backbones of our framework and optimized their weights with AdamW~\cite{loshchilov2018adamw}.
We set the batch size to 32, the learning rate to 1e-5, the warmup ratio to 0.1, and the weight decay to 0.01.
We set the loss scaling factor $\alpha$ to 0.1 for Llama and 0.4 for Phi to balance between unlearning and retention of information (see Appendix~\ref{app:retain_loss} for details).
We train for 5 epochs with early stopping applied.
All experiments were conducted with four NVIDIA H100 GPUs.
Each experiment was repeated with three different random seeds, and the results were averaged for reporting.

\paragraph{Knowledge unlearning approaches}

We evaluate the following state-of-the-art knowledge unlearning approaches (see Appendix~\ref{app:baseline_methods} for details):
\begin{itemize}
    \item \textbf{GA}~\cite{jang2023knowledgeunlearning}: Applies gradient ascent to decrease the likelihood of token sequences associated with the forget set
    \item \textbf{DPO}~\cite{rafailov2023dpo}: Performs direct preference optimization, prioritizing ``I don't know'' responses for items in the forget set
    \item \textbf{NPO}~\cite{zhang2024npo}: Implements negative preference optimization to actively disfavor responses linked to the forget set
    \item \textbf{+RT}: Includes additional finetuning on the retain set to explicitly reinforce knowledge retention in the model
    \vspace{-0.2cm}
\end{itemize}

\begin{figure*}
    \centering
    \includegraphics[width=\textwidth]{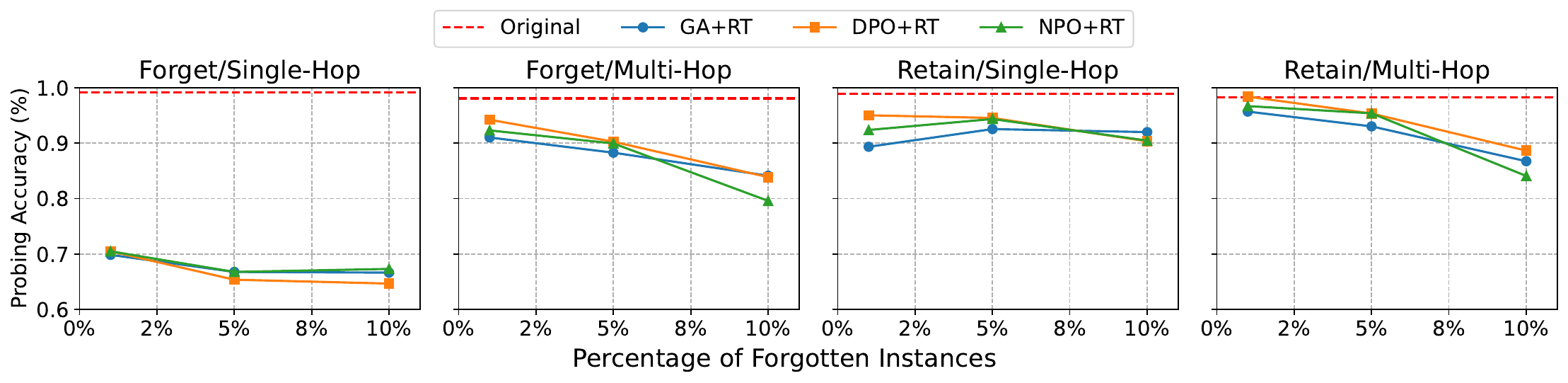}
    \caption{Scaling performance of various unlearning methods using Llama-3.1-8B-Instruct across different proportions of data for the forget set (1\%, 5\%, and 10\%). Models consistently preserve the ability to unlearn and retain single-hop facts with scaling. While unlearning multi-hop facts seems to improve with scaling, as evidenced by the performance drop, a similar decline is also observed in the retain set. This suggests that the effect may be attributed to catastrophic forgetting of broader information rather than a genuine improvement in unlearning multi-hop facts.}
    \label{fig:scaling_exp}
    % \vspace{-0.25cm}
\end{figure*}

\paragraph{Evaluation metrics}

To assess the unlearning of factual knowledge, we adopt the approach of \citet{petroni2019lama} and report \textbf{Probing Accuracy (PA)}.
This rank-based metric computes the mean precision at $k$ ($P@k$) across all relations, with $k$ set to 1.
In other words, for a given fact, the value is 1 if the correct object appears among the top $k$ predictions, and 0 otherwise.
By the definition of probing in Section~\ref{sec:probing}, we consider a pretrained language model to have successfully unlearned a fact if it can no longer predict the correct object accurately.
We also use \textbf{ROUGE-L recall (R-L)}~\cite{lin2004rouge} to compare the model's generated outputs (using greedy sampling) to the ground-truth answers.
This score serves as a proxy for accuracy in the question answering task, accounting for minor differences in phrasing between the generated and reference outputs.
Lastly, we measure the \textbf{Language Modeling Loss (LM)} over token sequences to determine how perplexed the model is by the data.

\begin{figure}[ht]
    \centering
    \includegraphics[width=\linewidth]{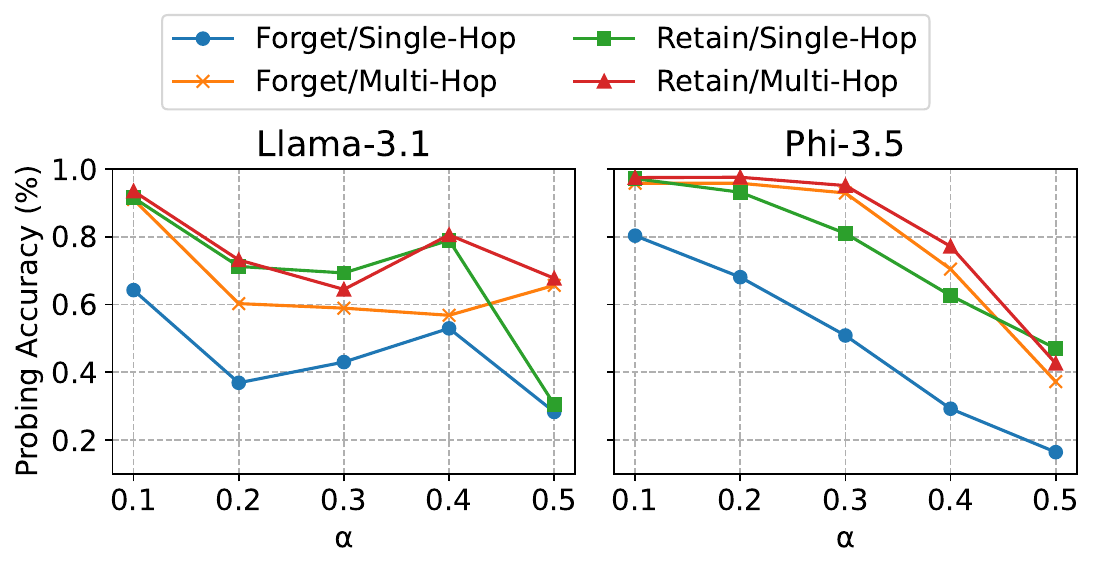}
    \caption{Performance of the GA+RT method with varying the loss scaling factor $\alpha$. Llama appears to be more sensitive than Phi to the value of $\alpha$ when balancing unlearning and retaining.}
    \label{fig:loss_scaling_exp}
    % \vspace{-0.25cm}
\end{figure}

\subsection{Knowledge Unlearning Results}

We present a comparison of unlearning performance across various methods in Table~\ref{tab:prelim_exp}.
Each method was trained for at least one epoch to ensure the model had exposure to all samples in the forget set.
Throughout the process, we prioritized model retention to maintain key metrics, such as PA and R-L, close to their original performance levels.
Our findings reveal that all unlearning methods (i.e., GA, DPO, and NPO) experienced model collapse after just one epoch, consistent with previous observations~\cite{lee2024pop}.
Whether factual knowledge was in the forget set or the retain set, models largely lost their ability to retain information and function correctly.
On the other hand, additional finetuning on the retain set (i.e., +RT) mitigated catastrophic forgetting, which is evidenced by retention performance comparable to the original for both single-hop and multi-hop facts.
Surprisingly, as shown in Figure~\ref{fig:loss_scaling_exp}, Llama appeared more sensitive to unlearning than Phi, requiring weaker unlearning scaling (i.e., $\alpha = 0.1$ for Llama versus $\alpha = 0.4$ for Phi), which explains the relatively high PA scores (e.g., $66.8\%$ with Llama vs. $34.9\%$ with Phi for GA+RT).
However, we also observe that while Phi maintained PA scores well with a higher $\alpha$ (even showing increases in PA for the retain set, likely due to Phi having been under-trained on the dataset), it exhibited poor retention in R-L scores (e.g., $81.2\% \rightarrow 58.2\%$ for NPO+RT).
We do not explore the variations of unlearning performance with different LLMs since they are not the focus of this work, but we hope these results can inspire future studies.
Lastly and most importantly, in both Llama and Phi models, multi-hop facts within the forget set were not effectively unlearned.
This indicates that existing unlearning methods, while capable of removing single-hop information, struggle to extend that effect to the corresponding multi-hop knowledge.
These outcomes underscore the need for new approaches to address unlearning in multi-hop scenarios.

\begin{figure*}
    \centering
    \includegraphics[width=\textwidth]{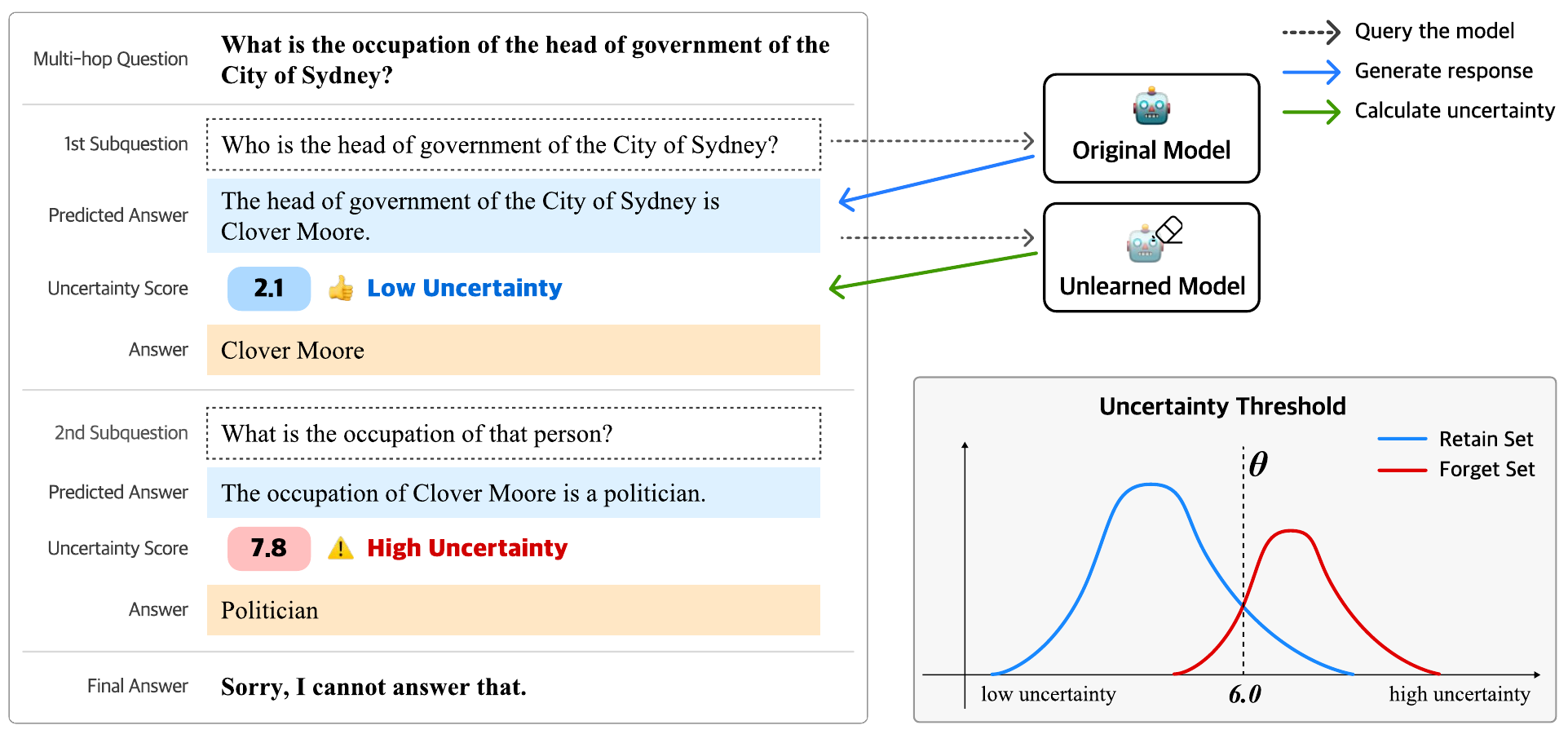}
    \caption{\textbf{Overview of the proposed \ours{} framework.} \ours{} begins by breaking down a multi-hop question into a sequence of subquestions, where each subquestion is passed to the original model to generate provisional answers. Then, \ours{} leverages the unlearned model to assess the uncertainty of each predicted answer by calculating uncertainty scores. If any subquestion yields a high uncertainty score -- exceeding a predefined threshold -- \ours{} responds with a rejection (e.g., ``I don't know''). Otherwise, the final response is based on the last intermediate answer in the sequence.}
    \label{fig:our_method}
    \vspace{-0.25cm}
\end{figure*}

\subsection{Evaluation with Unlearning at Scale}

In real-world scenarios, the number of samples to forget can vary. 
Thus, we evaluate the performance of unlearning and retaining multi-hop facts as the size of the forget set changes.
We conduct experiments using 1\%, 5\%, and 10\% of the dataset for forgetting (104, 523, and 1,046 single-hop instances, respectively), with the results shown in Figure~\ref{fig:scaling_exp}.
Our findings indicate that all knowledge unlearning methods effectively scale for single-hop, consistently preserving the ability to forget and retain single-hop facts.
Unlearning performance for single-hop facts stabilizes at approximately 66\%, while retention performance remains close to the original, around 90\%. 
For multi-hop facts, unlearning performance improves with larger forget sets, as reflected in a noticeable performance drop.
However, a similar decline is observed in the retain set, suggesting that this effect might stem from catastrophic forgetting of general knowledge rather than a true enhancement in unlearning multi-hop facts.

\section{\ours{}: A Proposal for Unlearning Multi-Hop Facts in LLMs}

In this section, we present \textbf{M}ulti-Hop unlearning via \textbf{UNC}ertainty t\textbf{H}reshold (\ours{}), a simple yet effective method to enhance the performance of unlearning multi-hop facts.
Figure~\ref{fig:our_method} illustrates the overview of our method.

% \usepackage{graphicx}
\begin{table*}[]
\small
\centering
% \resizebox{\textwidth}{!}{%
\begin{tabular}{l|cccc|cccc}
\toprule
 &
  \multicolumn{4}{c|}{\textit{Llama-3.1-8B-Instruct}} &
  \multicolumn{4}{c}{\textit{Phi-3.5-Mini-Instruct}} \\ \\[-2ex] \cline{2-9} \\[-2ex]
 &
  \multicolumn{2}{c}{\textbf{Forget Set}} &
  \multicolumn{2}{c|}{\textbf{Retain Set}} &
  \multicolumn{2}{c}{\textbf{Forget Set}} &
  \multicolumn{2}{c}{\textbf{Retain Set}} \\
 \multirow{-1}{*}{\textbf{Method}}
 &
  \textbf{PA}($\downarrow$) &
  \textbf{R-L}($\downarrow$) &
  \textbf{PA}($\uparrow$) &
  \textbf{R-L}($\uparrow$) &
  \textbf{PA}($\downarrow$) &
  \textbf{R-L}($\downarrow$) &
  \textbf{PA}($\uparrow$) &
  \textbf{R-L}($\uparrow$) \\ \midrule
Original &
  98.1 &
  29.2 &
  98.3 &
  28.0 &
  81.4 &
  52.6 &
  81.5 &
  53.8 \\
% \, w/ MeLLo\textsuperscript{\textdagger} (GPT-4o-mini) &
%   56.2 &
%   28.0 &
%   74.3 &
%   38.0 &
%   - &
%   - &
%   - &
%   - \\
\, w/ MeLLo\textsuperscript{\textdagger} &
  17.5 &
  6.7\textsuperscript{*} &
  84.8 &
  47.9\textsuperscript{*} &
  16.7 &
  6.7\textsuperscript{*} &
  81.9 &
  47.9\textsuperscript{*} \\ \midrule
GA+RT &
  88.3 &
  27.4 &
  93.0 &
  33.7 &
  80.9 &
  26.6 &
  85.3 &
  30.8 \\
\, w/ \ours{} &
  6.8 &
  2.2 &
  80.5 &
  29.3 &
  10.1 &
  3.1 &
  73.2 &
  26.7 \\ \midrule
DPO+RT &
  90.3 &
  24.0 &
  95.4 &
  33.3 &
  91.1 &
  6.7 &
  95.1 &
  9.8 \\
\, w/ \ours{} &
  6.7 &
  2.4 &
  80.8 &
  28.3 &
  11.0 &
  0.6 &
  82.1 &
  8.5 \\ \midrule
NPO+RT &
  90.0 &
  29.1 &
  95.4 &
  34.5 &
  88.5 &
  26.9 &
  90.2 &
  31.9 \\
\, w/ \ours{} &
  7.4 &
  2.8 &
  86.2 &
  31.6 &
  10.8 &
  3.8 &
  76.8 &
  27.3 \\ \bottomrule
\end{tabular}%
% }
\caption{Performance of multi-hop knowledge unlearning after applying post-hoc methods under the 5\% forget set setting. (\textdagger) indicates our modified implementation of MeLLo~\cite{zhong2023mquake}, adapted specifically for the unlearning task. It employs GPT-4o for both question decomposition and answer prediction. Scores marked with (*) indicate identical performance, as the predictions are derived directly from GPT-4o, while PA is computed using each model's respective probabilities. In contrast, \ours{} utilizes GPT-4o exclusively for question decomposition.}
% Our proposed method, \ours{}, is integrated into various knowledge unlearning approaches, significantly enhancing multi-hop knowledge unlearning performance across the board.}
\label{tab:posthoc_exp}
% \vspace{-0.25cm}
\end{table*}

\subsection{Methodology}

\paragraph{Decomposing multi-hop questions}

To enhance the unlearning of multi-hop facts in LLMs, we build on previous work by breaking down multi-hop questions into a series of simpler queries~\cite{zhou2023leasttomost}.
In multi-hop reasoning, where the predicted answer of one question serves as the subject for the next fact (i.e., $o_i = s_{i+1}$), model-generated responses to intermediate questions can slow down the process.
To mitigate this, we leverage coreference resolution to construct subquestions upfront, bypassing the need for sequential answering.
For instance, as shown in Figure~\ref{fig:our_method}, if the first subquestion is \textit{``Who is the head of government of the City of Sydney?''}, the second subquestion would be \textit{``What is the occupation of that person?''}, eliminating the need to resolve the first before proceeding.
We employ GPT-4o~\cite{achiam2023gpt4} to decompose the questions.

\paragraph{Predicting with original model}

After decomposing the multi-hop questions, each subquestion is fed into the model sequentially, along with the previously generated subquestions and their predicted answers.
It is important to note that the original model (i.e., the state prior to unlearning) is used to generate these answers, rather than the unlearned model.
As discussed in the next paragraph, we expect the uncertainty scores for outputs generated by the original model will be significantly high to the unlearned model, as these outputs resemble single-hop facts from the forget set.

\paragraph{Computing uncertainty scores}

Current knowledge unlearning methods generally reverse the language modeling objective to forget specific facts.
Therefore, we hypothesize that this approach will lead to high uncertainty in the unlearned model when presented with inputs resembling the single-hop facts from the forget set.
If a multi-hop question has effectively been decomposed, the resulting subquestions should closely resemble single-hop queries.
The outputs generated for these subquestions can therefore serve as proxies for the forgotten single-hop facts.
To measure uncertainty, we compute the negative log-likelihood of the token sequences generated by the original model, evaluated using the unlearned model.

\paragraph{Distinguishing forget and retain facts}

To determine which multi-hop facts to unlearn or retain, we establish a threshold that effectively separates the two data distributions.
To achieve this, we plot the probability density functions of the forget set and the validation split of the retain set, approximating the optimal threshold.
During inference, we apply this threshold to assess whether the uncertainty score of each subquestion's predicted answer is high or low.
If any subquestion yields a high uncertainty score, we replace the final answer with a rejective response (e.g., ``I don’t know''), following the approach of selective generation~\cite{zhang2024rtuning}.
Otherwise, the final answer is drawn from the last intermediate predicted answer in the multi-hop chain.
In practice, we first leverage external memory that stores single-hop facts from the forget set.
This allows us to bypass uncertainty calculations for subquestions that are highly similar to the stored facts using the Contriever retrieval model~\cite{izacard2022contriever}.
As the model size grows, this approach becomes increasingly efficient, as the retrieval model is significantly smaller and computationally lighter.

\begin{table*}[t!]

\centering
\begin{adjustbox}{width=\textwidth}
\begin{tabular}{ll}
\toprule
\textbf{Question} & Which continent is the country of origin of the Castres Olympique sport's team located in? (\textbf{Answer}: Europe)\\ \midrule
Original & The United Kingdom is located on the continent of \textcolor{red}{Europe}. \\
\,\, w/ MeLLo & \textcolor{red}{Europe} \\ \midrule
GA+RT & The country of origin of the (...) sport's team is France. The continent where the team is located is \textcolor{red}{Europe}. \\
\,\, w/ \textsc{Munch} & I must decline to answer due to \textbf{lack of information}. \\
 
\bottomrule
\end{tabular}
\end{adjustbox}
\vspace{-0.05in}
\caption{
Qualitative examples of generated sentences for a two-hop question, where the question contains a single-hop fact included in the forget set. The original answer, which is present in the model's response, is highlighted in \textcolor{red}{red}. The Llama-3.1-8B-Instruct model is used as the base model.
}
\label{tab:qualitative}
\vspace{-0.25cm}
\end{table*}

\subsection{Evaluation Results}

We incorporate \ours{} into previously unlearned models and present the results of multi-hop knowledge unlearning in Table~\ref{tab:posthoc_exp}.
To highlight the effectiveness of our approach, we compare it with MeLLo~\cite{zhong2023mquake}, a memory-based method designed for editing multi-hop knowledge.
Since MeLLo cannot be directly applied to unlearning tasks, we adapt the prompt so that the model generates a rejective response based on retrieval rather than editing the final answer.
Specifically, for each subquestion, the model generates a tentative answer and retrieves the most relevant fact, and if the two are equivalent, the model responds with a rejection because this signals that the model has produced an answer similar to one of the single-hop facts from the forget set.
For MeLLo, we use GPT-4o to decompose questions and predict answers, utilizing Llama or Phi only to calculate PA scores.
Table~\ref{tab:posthoc_exp} shows that while MeLLo improves unlearning performance, it slightly compromises retention.
Nevertheless, it achieves high R-L scores, likely due to the high-quality predictions from GPT-4o.
In contrast, \ours{} consistently outperforms MeLLo in multi-hop unlearning performance when integrated with any previously unlearned models (e.g., $98.1\% \rightarrow 6.8\%$ with Llama and $81.4\% \rightarrow 10.1\%$ with Phi for GA+RT).
While \ours{} also impacts retention to some extent, its ability to produce predictions directly from the models themselves (i.e., not using GPT-4o for predictions) offers promising performance with computational efficiency.

\subsection{Qualitative Analysis}

Table~\ref{tab:qualitative} presents qualitative results for a two-hop question.
The original model incorrectly identified the first hop's answer as ``United Kingdom'' instead of ``France,'' but still managed to produce the correct final answer ``Europe.''
MeLLo failed to abstain from generating the correct answer, likely due to relying heavily on the retriever model.
The GA+RT unlearned model also did not effectively erase multi-hop knowledge and arrived at the correct answer ``Europe'' through two-hop reasoning.
In contrast, our method \ours{} successfully abstained from answering correctly by leveraging a high uncertainty score.

\section{Related Work}

\subsection{Machine Unlearning}

Machine unlearning has gained prominence as a response to pressing issues surrounding data privacy, ethical considerations, and regulatory compliance in machine learning systems~\cite{cao2015mu, ginart2019clustering, bourtoule2021sisa}.
With the rise of LLMs, new methods have emerged to enable forgetting specific token sequences, such as through gradient ascent~\cite{jang2023knowledgeunlearning}, additional retention strategies~\cite{lee2024pop}, and negative preference optimization~\cite{zhang2024npo}.
However, \citet{choi2024crosslingual} identified a significant limitation in existing knowledge unlearning techniques: they fail to generalize across different languages, leaving models vulnerable to attacks in low-resource languages.
This reveals that when token sequences are substituted or aliased with alternative sequences, the unlearning process can be circumvented.
In this work, we investigate one of the indirect approaches to prompting models through multi-hop, which presents new challenges and opportunities for unlearning.

\subsection{Multi-Hop Reasoning}
Multi-hop reasoning involves deriving information by connecting multiple pieces of evidence across contexts \citep{huang-chang-2023-towards}. While recent LLMs excel at single-hop reasoning, their performance often declines with multi-hop reasoning that demands multiple steps and compositional reasoning \citep{wei2022emergent, zhou2023leasttomost}. This issue becomes especially challenging when LLMs need to update or delete knowledge, as it requires consistent propagation of changes across interconnected facts \citep{valmeekam2022large, press2023measuring, dziri2023faith, petty2024impact}. Most existing techniques for knowledge updating focus on modifying individual facts \citep{de2021editing, meng2022locating, zhang-etal-2024-knowledge-editing}, but often struggle with updating related knowledge \citep{onoe-etal-2023-lms, zhong2023mquake, cohen2024evaluating}. To resolve this, recent work has proposed solutions such as injecting information at inference time \citep{sakarvadia2023memory}, removing neurons responsible for shortcuts \citep{ju2024multihopshortcuts}, and inspecting model representations to fix multi-hop reasoning errors \citep{ghandeharioun2024patchscopes}. In contrast, our work examines whether removing specific information from models can be generalized effectively in multi-hop scenarios.

\section{Conclusion}

This study examines the effectiveness of current unlearning methods in eliminating multi-hop knowledge.
Our results indicate that they struggle to remove multi-hop knowledge when an intermediate hop is unlearned.
To overcome this issue, we propose \ours{}, a simple yet effective post-hoc approach that dissects multi-hop questions into simple subquestions and utilizes the unlearned model's uncertainty to decide when to provide a rejective response in the final decision-making process.
\ours{} significantly enhances the performance of multi-hop unlearning, offering a flexible and robust solution for unlearning in LLMs.

\section*{Limitations}

Our method, \ours{}, allows the model to refuse to answer multi-hop questions when uncertainty is high, drawing inspiration from selective generation~\cite{zhang2024rtuning}.
However, it does not essentially erase multi-hop knowledge from the model parameters, meaning this information could still be extracted through advanced adversarial techniques.
We recognize that attempting to remove multi-hop knowledge entirely through training could be risky, with potentially catastrophic consequences, as all knowledge might be interconnected through hops, raising the unresolved question of how many hops should be unlearned.
Additionally, \ours{} has room for improvement: using uncertainty scores as a decision-making threshold compromises the retention of some multi-hop knowledge.
We hope this work stimulates further research and discussions on creating a more robust framework for knowledge unlearning.

% \section*{Acknowledgments}

\bibliography{custom}

\appendix

\section{Additional Details for Knowledge Unlearning Methods}
\label{app:baseline_methods}

\subsection{GA}

Gradient ascent (GA)~\cite{jang2023knowledgeunlearning} reverses the language modeling loss, which can be understood as equivalent to gradient descent on the negative next-token prediction loss:
\begin{equation} \label{eq:1}
    \mathcal{L}_{\text{GA}} = - \mathbb{E}_{\mathcal{D}_f} [-\log (\pi_\theta (y|x))],
\end{equation}
which serves to minimize the token probabilities of the specific token sequences in the forget set $\mathcal{D}_f$.

\subsection{DPO}

In direct preference optimization (DPO)~\cite{rafailov2023dpo}, we are provided with a dataset of preference feedbacks $\mathcal{D}_{\text{paired}} = \{(x_i,y_{i,w},y_{i,l})\}_{i=1}^{N}$, where ``w'' stands for ``win'' and ``l'' stands for ``lose'' for two responses $y_w$ and $y_l$.
The goal is to train the model $\pi_\theta$ to align more closely with human preferences. In this work, the winning responses are rejections (e.g., ``I don't know.''), randomly sampled from 100 candidates used by \citet{maini2024tofu}.
Formally, DPO minimizes
\begin{equation} \label{eq:2}
    \begin{split}
        \mathcal{L}_{\text{DPO}} = - \mathbb{E}_{\mathcal{D}_{\text{paired}}} \left[ \log \sigma \left(\beta \log \frac{\pi_\theta(y_w|x)}{\pi_{\text{ref}}(y_w|x)} \right. \right. \\
        \left. \left. - \beta \log \frac{\pi_\theta(y_l|x)}{\pi_{\text{ref}}(y_l|x)} \right) \right],
    \end{split}
\end{equation}
where $\sigma(t)=1/(1+e^{-t})$ represents the sigmoid function, $\beta > 0$ is the inverse temperature, and $\phi_{\text{ref}}$ is a reference model.

\subsection{NPO}

Negative preference optimization (NPO)~\cite{zhang2024npo} ignores the $y_w$ term in DPO in Equation~\ref{eq:2} and aligns the language model with negative responses exclusively:
\begin{equation} \label{eq:3}
    \mathcal{L}_{\text{NPO}} = - \mathbb{E}_{\mathcal{D}_f} \left[ \log \sigma \left( -\beta \log \frac{\pi_\theta(y|x)}{\pi_{\text{ref}}(y|x)} \right) \right].
\end{equation}
Minimizing $\mathcal{L}_{\text{NPO}}$ drives the prediction probability $\pi_\theta(y|x)$ on the forget set to be as low as possible, effectively achieving the goal of unlearning the forget set.

\subsection{+RT} \label{app:retain_loss}

The explicit retention finetuning is achieved through standard language modeling on the retain set, which serves as the positive counterpart to Equation~\ref{eq:1}:
\begin{equation} \label{eq:4}
    \mathcal{L}_r = - \mathbb{E}_{\mathcal{D}_r} [\log (\pi_\theta (y|x))].
\end{equation}
Finally, the overall training objective is minimizing the following loss:
\begin{equation}
    \mathcal{L} = \alpha \cdot \mathcal{L}_f + (1-\alpha) \cdot \mathcal{L}_r,
\end{equation}
where $\mathcal{L}_f$ is one of the unlearning losses $\mathcal{L}_{\text{GA}}$, $\mathcal{L}_{\text{DPO}}$, or $\mathcal{L}_{\text{NPO}}$, and $\alpha$ is a loss scaling hyperparameter balancing the forgetting and retaining losses.

\section{Full Evaluation Results} \label{app:full_results}

In this work, we conducted experiments utilizing 1\%, 5\%, and 10\% of data as the forget set, while the rest of the dataset is used for the retain set.
We additionally report the 1\% and 10\% settings in Table~\ref{tab:prelim_exp_01} and Table~\ref{tab:prelim_exp_10}.
Post-hoc results are provided in Table~\ref{tab:posthoc_exp_01} and Table~\ref{tab:posthoc_exp_10}.
The observed trends align with those discussed in the main body of this paper.
Furthermore, we demonstrate model utility performance for the 1\%, 5\%, and 10\% settings in Tables~\ref{tab:utility_01}, \ref{tab:utility_05}, and \ref{tab:utility_10}.
To assess model utility, we validate our framework on eight language understanding benchmarks including ARC-Challenge~\cite{clark2018arc}, CommonsenseQA~\cite{talmor2019commonsenseqa}, HellaSwag~\cite{zellers2019hellaswag}, Lambada~\cite{paperno2016lambada}, MMLU~\cite{hendrycks2021mmlu}, OpenbookQA~\cite{mihaylov2018obqa}, PIQA~\cite{bisk2020piqa}, and Winogrande~\cite{sakaguchi2021winogrande}.

% \usepackage{graphicx}
\begin{table*}[]
\small
\centering
\resizebox{\textwidth}{!}{%
\begin{tabular}{lcccccccccccc}
\toprule
\multicolumn{1}{l|}{} &
  \multicolumn{3}{c|}{\textbf{Forget Set (Single-Hop)}} &
  \multicolumn{3}{c|}{\textbf{Forget Set (Multi-Hop)}} &
  \multicolumn{3}{c|}{\textbf{Retain Set (Single-Hop)}} &
  \multicolumn{3}{c}{\textbf{Retain Set (Multi-Hop)}} \\
\multicolumn{1}{l|}{} &
  \textbf{PA}($\downarrow$) &
  \textbf{R-L}($\downarrow$) &
  \multicolumn{1}{c|}{\textbf{LM}($\uparrow$)} &
  \textbf{PA}($\downarrow$) &
  \textbf{R-L}($\downarrow$) &
  \multicolumn{1}{c|}{\textbf{LM}($\uparrow$)} &
  \textbf{PA}($\uparrow$) &
  \textbf{R-L}($\uparrow$) &
  \multicolumn{1}{c|}{\textbf{LM}($\downarrow$)} &
  \textbf{PA}($\uparrow$) &
  \textbf{R-L}($\uparrow$) &
  \textbf{LM}($\downarrow$) \\ \\[-2ex]\hline
\multicolumn{13}{l}{\cellcolor[HTML]{EFEFEF}\textit{Llama-3.1-8B-Instruct}} \\ \hline \\[-2ex]
\multicolumn{1}{l|}{Original} &
  99.0 &
  61.0 &
  \multicolumn{1}{c|}{0.6} &
  99.0 &
  31.1 &
  \multicolumn{1}{c|}{1.0} &
  98.9 &
  60.0 &
  \multicolumn{1}{c|}{0.6} &
  98.3 &
  28.0 &
  1.0 \\ \midrule
\multicolumn{1}{l|}{GA\textsuperscript{*}} &
  15.7 &
  0.0 &
  \multicolumn{1}{c|}{107.4} &
  26.0 &
  0.0 &
  \multicolumn{1}{c|}{106.7} &
  21.2 &
  0.0 &
  \multicolumn{1}{c|}{107.2} &
  25.2 &
  0.0 &
  106.6 \\
\multicolumn{1}{l|}{DPO\textsuperscript{*}} &
  25.0 &
  1.5 &
  \multicolumn{1}{c|}{91.6} &
  27.2 &
  0.5 &
  \multicolumn{1}{c|}{90.8} &
  24.4 &
  1.0 &
  \multicolumn{1}{c|}{91.5} &
  21.0 &
  0.5 &
  90.7 \\
\multicolumn{1}{l|}{NPO\textsuperscript{*}} &
  25.0 &
  0.0 &
  \multicolumn{1}{c|}{102.4} &
  41.0 &
  0.0 &
  \multicolumn{1}{c|}{101.9} &
  26.7 &
  0.0 &
  \multicolumn{1}{c|}{102.3} &
  38.6 &
  0.0 &
  101.9 \\ \midrule
\multicolumn{1}{l|}{GA+RT} &
  \textbf{69.9} &
  49.1 &
  \multicolumn{1}{c|}{\textbf{7.5}} &
  \textbf{91.0} &
  26.0 &
  \multicolumn{1}{c|}{\textbf{4.4}} &
  89.4 &
  75.5 &
  \multicolumn{1}{c|}{3.1} &
  95.7 &
  28.9 &
  3.8 \\
\multicolumn{1}{l|}{DPO+RT} &
  70.5 &
  \textbf{26.9} &
  \multicolumn{1}{c|}{4.9} &
  94.2 &
  \textbf{18.6} &
  \multicolumn{1}{c|}{2.2} &
  \textbf{95.0} &
  58.6 &
  \multicolumn{1}{c|}{\textbf{1.4}} &
  \textbf{98.4} &
  23.6 &
  \textbf{1.7} \\
\multicolumn{1}{l|}{NPO+RT} &
  70.5 &
  51.7 &
  \multicolumn{1}{c|}{6.3} &
  92.3 &
  24.9 &
  \multicolumn{1}{c|}{3.3} &
  92.4 &
  \textbf{77.3} &
  \multicolumn{1}{c|}{2.4} &
  96.7 &
  \textbf{31.2} &
  2.7 \\ \\[-2ex]\hline \hline
\multicolumn{13}{l}{\cellcolor[HTML]{EFEFEF}\textit{Phi-3.5-Mini-Instruct}} \\ \hline \\[-2ex]
\multicolumn{1}{l|}{Original} &
  83.7 &
  85.0 &
  \multicolumn{1}{c|}{3.8} &
  78.8 &
  55.7 &
  \multicolumn{1}{c|}{3.3} &
  87.7 &
  81.2 &
  \multicolumn{1}{c|}{3.7} &
  81.5 &
  53.8 &
  3.4 \\ \midrule
\multicolumn{1}{l|}{GA\textsuperscript{*}} &
  17.6 &
  0.1 &
  \multicolumn{1}{c|}{129.4} &
  25.3 &
  0.0 &
  \multicolumn{1}{c|}{126.8} &
  16.1 &
  0.2 &
  \multicolumn{1}{c|}{129.4} &
  21.9 &
  0.0 &
  126.8 \\
\multicolumn{1}{l|}{DPO\textsuperscript{*}} &
  10.6 &
  0.0 &
  \multicolumn{1}{c|}{87.2} &
  18.9 &
  0.0 &
  \multicolumn{1}{c|}{82.5} &
  14.5 &
  0.0 &
  \multicolumn{1}{c|}{86.6} &
  16.9 &
  0.1 &
  82.3 \\
\multicolumn{1}{l|}{NPO\textsuperscript{*}} &
  16.0 &
  0.2 &
  \multicolumn{1}{c|}{110.5} &
  28.5 &
  0.2 &
  \multicolumn{1}{c|}{108.7} &
  16.0 &
  0.7 &
  \multicolumn{1}{c|}{110.6} &
  23.6 &
  0.2 &
  109.0 \\ \midrule
\multicolumn{1}{l|}{GA+RT} &
  47.4 &
  46.5 &
  \multicolumn{1}{c|}{6.3} &
  90.4 &
  34.8 &
  \multicolumn{1}{c|}{2.6} &
  82.3 &
  66.8 &
  \multicolumn{1}{c|}{2.4} &
  \textbf{90.7} &
  39.4 &
  2.3 \\
\multicolumn{1}{l|}{DPO+RT} &
  55.1 &
  \textbf{8.6} &
  \multicolumn{1}{c|}{4.4} &
  \textbf{84.6} &
  \textbf{4.6} &
  \multicolumn{1}{c|}{2.0} &
  \textbf{85.3} &
  34.8 &
  \multicolumn{1}{c|}{\textbf{1.6}} &
  89.9 &
  8.2 &
  \textbf{1.6} \\
\multicolumn{1}{l|}{NPO+RT} &
  \textbf{43.6} &
  46.4 &
  \multicolumn{1}{c|}{\textbf{7.1}} &
  87.5 &
  36.1 &
  \multicolumn{1}{c|}{\textbf{3.1}} &
  80.6 &
  \textbf{67.5} &
  \multicolumn{1}{c|}{2.8} &
  88.7 &
  \textbf{40.8} &
  2.7 \\ \bottomrule
\end{tabular}%
}
\caption{Performance comparison of different knowledge unlearning methods after erasing single-hop facts from the forget set in Llama-3.1-8B-Instruct and Phi-3.5-Mini-Instruct models, utilizing 1\% of data as the forget set. Models marked with (*) indicate collapse. The best results are highlighted in \textbf{bold}.}
\label{tab:prelim_exp_01}
\end{table*}
% \usepackage{graphicx}
\begin{table*}[]
\small
\centering
\resizebox{\textwidth}{!}{%
\begin{tabular}{lcccccccccccc}
\toprule
\multicolumn{1}{l|}{} &
  \multicolumn{3}{c|}{\textbf{Forget Set (Single-Hop)}} &
  \multicolumn{3}{c|}{\textbf{Forget Set (Multi-Hop)}} &
  \multicolumn{3}{c|}{\textbf{Retain Set (Single-Hop)}} &
  \multicolumn{3}{c}{\textbf{Retain Set (Multi-Hop)}} \\
\multicolumn{1}{l|}{} &
  \textbf{PA}($\downarrow$) &
  \textbf{R-L}($\downarrow$) &
  \multicolumn{1}{c|}{\textbf{LM}($\uparrow$)} &
  \textbf{PA}($\downarrow$) &
  \textbf{R-L}($\downarrow$) &
  \multicolumn{1}{c|}{\textbf{LM}($\uparrow$)} &
  \textbf{PA}($\uparrow$) &
  \textbf{R-L}($\uparrow$) &
  \multicolumn{1}{c|}{\textbf{LM}($\downarrow$)} &
  \textbf{PA}($\uparrow$) &
  \textbf{R-L}($\uparrow$) &
  \textbf{LM}($\downarrow$) \\ \\[-2ex]\hline
\multicolumn{13}{l}{\cellcolor[HTML]{EFEFEF}\textit{Llama-3.1-8B-Instruct}} \\ \hline \\[-2ex]
\multicolumn{1}{l|}{Original} &
  99.1 &
  58.3 &
  \multicolumn{1}{c|}{0.6} &
  98.1 &
  29.1 &
  \multicolumn{1}{c|}{1.1} &
  98.9 &
  60.0 &
  \multicolumn{1}{c|}{0.6} &
  98.3 &
  28.0 &
  1.0 \\ \midrule
\multicolumn{1}{l|}{GA\textsuperscript{*}} &
  6.8 &
  0.1 &
  \multicolumn{1}{c|}{192.6} &
  9.9 &
  0.0 &
  \multicolumn{1}{c|}{189.2} &
  8.1 &
  0.0 &
  \multicolumn{1}{c|}{192.1} &
  10.5 &
  0.0 &
  189.2 \\
\multicolumn{1}{l|}{DPO\textsuperscript{*}} &
  26.2 &
  0.5 &
  \multicolumn{1}{c|}{117.1} &
  28.7 &
  0.4 &
  \multicolumn{1}{c|}{115.3} &
  27.7 &
  0.5 &
  \multicolumn{1}{c|}{117.0} &
  29.7 &
  0.2 &
  115.1 \\
\multicolumn{1}{l|}{NPO\textsuperscript{*}} &
  8.2 &
  0.1 &
  \multicolumn{1}{c|}{119.1} &
  18.0 &
  0.0 &
  \multicolumn{1}{c|}{117.6} &
  9.2 &
  0.0 &
  \multicolumn{1}{c|}{118.9} &
  17.7 &
  0.0 &
  117.6 \\ \midrule
\multicolumn{1}{l|}{GA+RT} &
  66.7 &
  49.2 &
  \multicolumn{1}{c|}{12.3} &
  84.1 &
  26.7 &
  \multicolumn{1}{c|}{7.9} &
  \textbf{92.0} &
  \textbf{71.3} &
  \multicolumn{1}{c|}{4.7} &
  86.7 &
  \textbf{31.7} &
  6.2 \\
\multicolumn{1}{l|}{DPO+RT} &
  \textbf{64.7} &
  \textbf{25.8} &
  \multicolumn{1}{c|}{11.1} &
  83.9 &
  \textbf{18.8} &
  \multicolumn{1}{c|}{5.8} &
  90.3 &
  56.0 &
  \multicolumn{1}{c|}{\textbf{4.3}} &
  \textbf{88.7} &
  26.6 &
  \textbf{4.5} \\
\multicolumn{1}{l|}{NPO+RT} &
  67.3 &
  50.4 &
  \multicolumn{1}{c|}{\textbf{12.8}} &
  \textbf{79.6} &
  25.7 &
  \multicolumn{1}{c|}{\textbf{9.4}} &
  90.5 &
  71.2 &
  \multicolumn{1}{c|}{4.4} &
  84.1 &
  30.5 &
  7.6 \\ \\[-2ex]\hline \hline
\multicolumn{13}{l}{\cellcolor[HTML]{EFEFEF}\textit{Phi-3.5-Mini-Instruct}} \\ \hline \\[-2ex]
\multicolumn{1}{l|}{Original} &
  88.8 &
  79.9 &
  \multicolumn{1}{c|}{3.7} &
  81.9 &
  51.7 &
  \multicolumn{1}{c|}{3.5} &
  87.7 &
  81.2 &
  \multicolumn{1}{c|}{3.7} &
  81.5 &
  53.8 &
  3.4 \\ \midrule
\multicolumn{1}{l|}{GA\textsuperscript{*}} &
  1.1 &
  0.1 &
  \multicolumn{1}{c|}{220.1} &
  8.2 &
  0.0 &
  \multicolumn{1}{c|}{205.7} &
  2.6 &
  0.0 &
  \multicolumn{1}{c|}{218.6} &
  8.0 &
  0.0 &
  205.4 \\
\multicolumn{1}{l|}{DPO\textsuperscript{*}} &
  11.3 &
  0.5 &
  \multicolumn{1}{c|}{123.0} &
  15.1 &
  0.2 &
  \multicolumn{1}{c|}{121.3} &
  11.5 &
  0.3 &
  \multicolumn{1}{c|}{122.9} &
  14.6 &
  0.2 &
  121.0 \\
\multicolumn{1}{l|}{NPO\textsuperscript{*}} &
  24.4 &
  0.1 &
  \multicolumn{1}{c|}{145.9} &
  22.5 &
  0.0 &
  \multicolumn{1}{c|}{143.4} &
  22.4 &
  0.0 &
  \multicolumn{1}{c|}{145.8} &
  21.5 &
  0.0 &
  143.2 \\ \midrule
\multicolumn{1}{l|}{GA+RT} &
  \textbf{50.4} &
  24.8 &
  \multicolumn{1}{c|}{\textbf{35.9}} &
  \textbf{72.7} &
  13.7 &
  \multicolumn{1}{c|}{\textbf{19.3}} &
  81.3 &
  36.2 &
  \multicolumn{1}{c|}{16.1} &
  80.1 &
  16.1 &
  13.5 \\
\multicolumn{1}{l|}{DPO+RT} &
  55.6 &
  \textbf{7.8} &
  \multicolumn{1}{c|}{14.3} &
  85.2 &
  \textbf{5.5} &
  \multicolumn{1}{c|}{4.9} &
  \textbf{89.4} &
  28.3 &
  \multicolumn{1}{c|}{\textbf{4.3}} &
  \textbf{89.2} &
  7.1 &
  \textbf{3.7} \\
\multicolumn{1}{l|}{NPO+RT} &
  52.1 &
  36.8 &
  \multicolumn{1}{c|}{23.7} &
  83.0 &
  21.0 &
  \multicolumn{1}{c|}{6.3} &
  85.8 &
  \textbf{55.5} &
  \multicolumn{1}{c|}{7.5} &
  88.4 &
  \textbf{24.3} &
  4.3 \\ \bottomrule
\end{tabular}%
}
\caption{Performance comparison of different knowledge unlearning methods after erasing single-hop facts from the forget set in Llama-3.1-8B-Instruct and Phi-3.5-Mini-Instruct models, utilizing 10\% of data as the forget set. Models marked with (*) indicate collapse. The best results are highlighted in \textbf{bold}.}
\label{tab:prelim_exp_10}
\end{table*}
% \usepackage{graphicx}
\begin{table*}[ht]
\small
\centering
% \resizebox{\textwidth}{!}{%
\begin{tabular}{l|cccc|cccc}
\toprule
 &
  \multicolumn{4}{c|}{\textit{Llama-3.1-8B-Instruct}} &
  \multicolumn{4}{c}{\textit{Phi-3.5-Mini-Instruct}} \\ \\[-2ex] \cline{2-9} \\[-2ex]
 &
  \multicolumn{2}{c}{\textbf{Forget Set}} &
  \multicolumn{2}{c|}{\textbf{Retain Set}} &
  \multicolumn{2}{c}{\textbf{Forget Set}} &
  \multicolumn{2}{c}{\textbf{Retain Set}} \\
 \multirow{-1}{*}{\textbf{Method}}
 &
  \textbf{PA}($\downarrow$) &
  \textbf{R-L}($\downarrow$) &
  \textbf{PA}($\uparrow$) &
  \textbf{R-L}($\uparrow$) &
  \textbf{PA}($\downarrow$) &
  \textbf{R-L}($\downarrow$) &
  \textbf{PA}($\uparrow$) &
  \textbf{R-L}($\uparrow$) \\ \midrule
Original &
  99.0 &
  31.1 &
  98.3 &
  28.0 &
  78.8 &
  55.7 &
  81.5 &
  53.8 \\
\, w/ MeLLo\textsuperscript{\textdagger} &
  14.4 &
  6.3\textsuperscript{*} &
  84.8 &
  47.9\textsuperscript{*} &
  14.4 &
  6.3\textsuperscript{*} &
  81.9 &
  47.9\textsuperscript{*} \\ \midrule
GA+RT &
  91.0 &
  26.0 &
  95.7 &
  28.9 &
  90.4 &
  34.8 &
  90.7 &
  39.4 \\
\, w/ \ours{} &
  7.1 &
  4.1 &
  71.8 &
  21.1 &
  11.5 &
  4.1 &
  79.4 &
  35.5 \\ \midrule
DPO+RT &
  94.2 &
  18.6 &
  98.4 &
  23.6 &
  84.6 &
  4.6 &
  89.9 &
  8.2 \\
\, w/ \ours{} &
  7.4 &
  3.1 &
  83.8 &
  19.9 &
  7.1 &
  0.1 &
  73.3 &
  7.7 \\ \midrule
NPO+RT &
  92.3 &
  24.9 &
  96.7 &
  31.2 &
  87.5 &
  36.1 &
  88.7 &
  40.8 \\
\, w/ \ours{} &
  5.8 &
  4.2 &
  75.1 &
  24.4 &
  11.5 &
  4.4 &
  78.6 &
  37.1 \\ \bottomrule
\end{tabular}%
% }
\caption{Performance of multi-hop knowledge unlearning after applying post-hoc methods under the 1\% forget set setting. (\textdagger) indicates our modified implementation of MeLLo, adapted specifically for the unlearning task. Scores marked with (*) indicate identical performance, as the predictions are derived directly from GPT-4o.}
\label{tab:posthoc_exp_01}
\end{table*}
% \usepackage{graphicx}
\begin{table*}[ht]
\small
\centering
% \resizebox{\textwidth}{!}{%
\begin{tabular}{l|cccc|cccc}
\toprule
 & \multicolumn{4}{c|}{\textit{Llama-3.1-8B-Instruct}} & 
   \multicolumn{4}{c}{\textit{Phi-3.5-Mini-Instruct}} \\ \\[-2ex] \cline{2-9} \\[-2ex]
 &
  \multicolumn{2}{c}{\textbf{Forget Set}} &
  \multicolumn{2}{c|}{\textbf{Retain Set}} &
  \multicolumn{2}{c}{\textbf{Forget Set}} &
  \multicolumn{2}{c}{\textbf{Retain Set}} \\
 \multirow{-1}{*}{\textbf{Method}}
&
  \textbf{PA}($\downarrow$) &
  \textbf{R-L}($\downarrow$) &
  \textbf{PA}($\uparrow$) &
  \textbf{R-L}($\uparrow$) &
  \textbf{PA}($\downarrow$) &
  \textbf{R-L}($\downarrow$) &
  \textbf{PA}($\uparrow$) &
  \textbf{R-L}($\uparrow$) \\ \midrule
Original &
  98.1 & 
  29.1 & 
  98.3 & 
  28.0 & 
  81.9 &
  51.7 & 
  81.5 &
  53.8 \\
\, w/ MeLLo\textsuperscript{\textdagger} & 
  17.5 &
  7.9\textsuperscript{*} & 
  84.8 &
  47.9\textsuperscript{*} &
  17.1 & 
  7.9\textsuperscript{*} &
  81.9 & 
  47.9\textsuperscript{*} \\ \midrule
GA+RT & 
  84.1 & 
  26.7 &
  86.7 & 
  31.7 &
  72.7 &
  13.7 &
  80.1 &
  16.1 \\
\, w/ \ours{} &
  5.9 &
  2.7 & 
  69.7 & 
  26.3 &
  7.8  & 
  2.1 &
  70.0 & 
  14.2 \\ \midrule
DPO+RT &
  83.9 & 
  18.8 & 
  88.7 & 
  26.6 & 
  85.2 & 
  5.5 & 
  89.2 & 
  7.1 \\
\, w/ \ours{} &
  5.8 & 
  2.0 & 
  70.8 & 
  22.8 & 
  8.7 & 
  1.1 & 
  76.1 & 
  6.4 \\ \midrule
NPO+RT & 
  79.6 & 
  25.7 &
  84.1 & 
  30.5 & 
  83.0 &
  21.0 & 
  88.4 &
  24.3 \\
\, w/ \ours{} &
  5.8 & 
  2.6 &
  67.3 & 
  26.0 & 
  8.4 & 
  2.9 & 
  76.3 & 
  21.5 \\ \bottomrule
\end{tabular}%
\caption{Performance of multi-hop knowledge unlearning after applying post-hoc methods under the 10\% forget set setting. (\textdagger) indicates our modified implementation of MeLLo, adapted specifically for the unlearning task. Scores marked with (*) indicate identical performance, as the predictions are derived directly from GPT-4o.}
\label{tab:posthoc_exp_10}
\end{table*}
\clearpage
% \usepackage{graphicx}
\begin{table*}[]
\small
\centering
% \resizebox{\textwidth}{!}{%
\begin{tabular}{lccccccccc}
\toprule
\multicolumn{1}{l|}{} &
  \textbf{ARC-C} &
  \textbf{CSQA} &
  \textbf{Hella.} &
  \textbf{Lamba.} &
  \textbf{MMLU} &
  \textbf{OBQA} &
  \textbf{PIQA} &
  \multicolumn{1}{c|}{\textbf{Wino.}} &
  \textbf{Avg.} \\ \\[-2ex] \hline
\multicolumn{10}{l}{\cellcolor[HTML]{EFEFEF}\textit{Llama-3.1-8B-Instruct}}                                       \\ \hline \\[-2ex]
\multicolumn{1}{l|}{Original} & 51.8 & 77.1 & 59.2 & 73.2 & 68.1 & 33.8 & 80.2 & \multicolumn{1}{c|}{74.1} & 64.7 \\ \midrule
\multicolumn{1}{l|}{GA+RT}    & 48.4 & 73.2 & 58.1 & 77.5 & 64.9 & 35.5 & 78.6 & \multicolumn{1}{c|}{72.6} & 63.6 \\
\multicolumn{1}{l|}{DPO+RT}   & 47.1 & 73.7 & 57.8 & 73.4 & 65.3 & 34.3 & 78.9 & \multicolumn{1}{c|}{72.8} & 62.9 \\
\multicolumn{1}{l|}{NPO+RT}   & 47.4 & 72.8 & 58.1 & 76.9 & 65.1 & 35.3 & 78.6 & \multicolumn{1}{c|}{72.8} & 63.4 \\ \\[-2ex] \hline \hline
\multicolumn{10}{l}{\cellcolor[HTML]{EFEFEF}\textit{Phi-3.5-Mini-Instruct}}                                       \\ \hline \\[-2ex]
\multicolumn{1}{l|}{Original} & 59.5 & 75.3 & 58.8 & 65.1 & 68.7 & 37.6 & 80.0 & \multicolumn{1}{c|}{74.6} & 65.0 \\ \midrule
\multicolumn{1}{l|}{GA+RT}    & 59.2 & 75.2 & 59.5 & 63.4 & 68.7 & 39.3 & 79.1 & \multicolumn{1}{c|}{74.0} & 64.8 \\
\multicolumn{1}{l|}{DPO+RT}   & 59.8 & 74.5 & 58.0 & 60.2 & 68.4 & 38.5 & 80.3 & \multicolumn{1}{c|}{76.4} & 64.5 \\
\multicolumn{1}{l|}{NPO+RT}   & 59.1 & 75.2 & 59.6 & 63.4 & 68.6 & 39.1 & 79.1 & \multicolumn{1}{c|}{73.7} & 64.7 \\ \bottomrule
\end{tabular}%
% }
\caption{Model utility performance after erasing single-hop facts from the forget set in Llama-3.1-8B-Instruct and Phi-3.5-Mini-Instruct models, utilizing 1\% of data as the forget set.}
\label{tab:utility_01}
\end{table*}
% \usepackage{graphicx}
\begin{table*}[]
\small
\centering
% \resizebox{\textwidth}{!}{%
\begin{tabular}{lccccccccc}
\toprule
\multicolumn{1}{l|}{} &
  \textbf{ARC-C} &
  \textbf{CSQA} &
  \textbf{Hella.} &
  \textbf{Lamba.} &
  \textbf{MMLU} &
  \textbf{OBQA} &
  \textbf{PIQA} &
  \multicolumn{1}{c|}{\textbf{Wino.}} &
  \textbf{Avg.} \\ \\[-2ex] \hline
\multicolumn{10}{l}{\cellcolor[HTML]{EFEFEF}\textit{Llama-3.1-8B-Instruct}}                                       \\ \hline \\[-2ex]
\multicolumn{1}{l|}{Original} & 51.8 & 77.1 & 59.2 & 73.2 & 68.1 & 33.8 & 80.2 & \multicolumn{1}{c|}{74.1} & 64.7 \\ \midrule
\multicolumn{1}{l|}{GA+RT}    & 42.9 & 59.3 & 56.0 & 79.5 & 63.7 & 35.3 & 76.3 & \multicolumn{1}{c|}{72.1} & 60.6 \\
\multicolumn{1}{l|}{DPO+RT}   & 45.9 & 67.6 & 56.3 & 74.9 & 63.8 & 35.8 & 77.9 & \multicolumn{1}{c|}{71.4} & 61.7 \\
\multicolumn{1}{l|}{NPO+RT}   & 46.6 & 61.3 & 56.7 & 78.4 & 64.4 & 35.7 & 78.0 & \multicolumn{1}{c|}{72.2} & 61.7 \\ \\[-2ex] \hline \hline
\multicolumn{10}{l}{\cellcolor[HTML]{EFEFEF}\textit{Phi-3.5-Mini-Instruct}}                                       \\ \hline \\[-2ex]
\multicolumn{1}{l|}{Original} & 59.5 & 75.3 & 58.8 & 65.1 & 68.7 & 37.6 & 80.0 & \multicolumn{1}{c|}{74.6} & 65.0 \\ \midrule
\multicolumn{1}{l|}{GA+RT}    & 61.1 & 74.9 & 59.6 & 67.6 & 68.6 & 40.1 & 78.8 & \multicolumn{1}{c|}{73.5} & 65.5 \\
\multicolumn{1}{l|}{DPO+RT}   & 59.0 & 74.6 & 57.7 & 63.0 & 68.0 & 39.0 & 79.5 & \multicolumn{1}{c|}{74.8} & 64.5 \\
\multicolumn{1}{l|}{NPO+RT}   & 60.1 & 75.1 & 59.7 & 67.7 & 68.6 & 40.1 & 78.9 & \multicolumn{1}{c|}{73.6} & 65.5 \\ \bottomrule
\end{tabular}%
% }
\caption{Model utility performance after erasing single-hop facts from the forget set in Llama-3.1-8B-Instruct and Phi-3.5-Mini-Instruct models, utilizing 5\% of data as the forget set.}
\label{tab:utility_05}
\end{table*}
% \usepackage{graphicx}
\begin{table*}[]
\small
\centering
% \resizebox{\textwidth}{!}{%
\begin{tabular}{lccccccccc}
\toprule
\multicolumn{1}{l|}{} &
  \textbf{ARC-C} &
  \textbf{CSQA} &
  \textbf{Hella.} &
  \textbf{Lamba.} &
  \textbf{MMLU} &
  \textbf{OBQA} &
  \textbf{PIQA} &
  \multicolumn{1}{c|}{\textbf{Wino.}} &
  \textbf{Avg.} \\ \\[-2ex] \hline
\multicolumn{10}{l}{\cellcolor[HTML]{EFEFEF}\textit{Llama-3.1-8B-Instruct}}                                       \\ \hline \\[-2ex]
\multicolumn{1}{l|}{Original} & 51.8 & 77.1 & 59.2 & 73.2 & 68.1 & 33.8 & 80.2 & \multicolumn{1}{c|}{74.1} & 64.7 \\ \midrule
\multicolumn{1}{l|}{GA+RT}    & 44.7 & 60.9 & 55.7 & 77.1 & 62.3 & 36.5 & 76.3 & \multicolumn{1}{c|}{71.7} & 60.6 \\
\multicolumn{1}{l|}{DPO+RT}   & 45.4 & 66.9 & 56.2 & 75.5 & 62.8 & 34.9 & 77.0 & \multicolumn{1}{c|}{70.2} & 61.1 \\
\multicolumn{1}{l|}{NPO+RT}   & 45.6 & 64.5 & 56.6 & 78.6 & 62.5 & 36.5 & 77.9 & \multicolumn{1}{c|}{72.3} & 61.8 \\ \\[-2ex] \hline \hline
\multicolumn{10}{l}{\cellcolor[HTML]{EFEFEF}\textit{Phi-3.5-Mini-Instruct}}                                       \\ \hline \\[-2ex]
\multicolumn{1}{l|}{Original} & 59.5 & 75.3 & 58.8 & 65.1 & 68.7 & 37.6 & 80.0 & \multicolumn{1}{c|}{74.6} & 65.0 \\ \midrule
\multicolumn{1}{l|}{GA+RT}    & 59.2 & 74.7 & 60.2 & 67.5 & 68.5 & 40.2 & 77.6 & \multicolumn{1}{c|}{71.6} & 64.9 \\
\multicolumn{1}{l|}{DPO+RT}   & 59.5 & 73.8 & 58.8 & 63.3 & 68.4 & 39.2 & 78.6 & \multicolumn{1}{c|}{72.3} & 64.2 \\
\multicolumn{1}{l|}{NPO+RT}   & 59.6 & 74.9 & 59.9 & 67.7 & 68.4 & 39.7 & 78.4 & \multicolumn{1}{c|}{71.5} & 65.0 \\ \bottomrule
\end{tabular}%
% }
\caption{Model utility performance after erasing single-hop facts from the forget set in Llama-3.1-8B-Instruct and Phi-3.5-Mini-Instruct models, utilizing 10\% of data as the forget set.}
\label{tab:utility_10}
\end{table*}

\begin{figure*}[h]
    \centering
    \includegraphics[width=\textwidth]{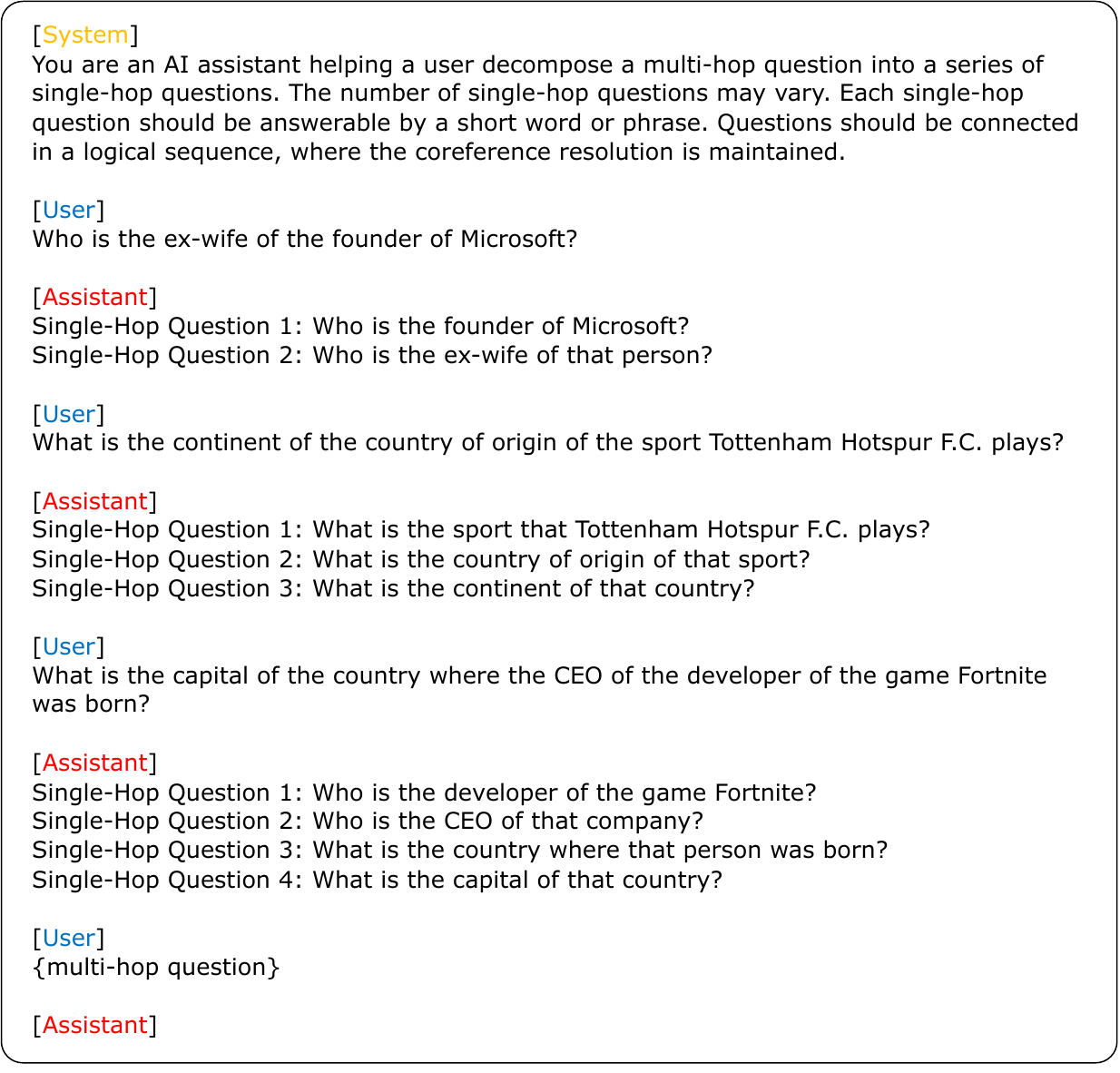}
    \caption{Prompt used in \ours{} to decompose a multi-hop question into a series of subquestions using GPT-4o. It consists of a system prompt followed by three fixed demonstration examples.}
    \label{fig:munch_prompt}
\end{figure*}

\end{document}